\documentclass{article}
\usepackage{ijcai24}

\usepackage{times}
\usepackage{soul}
\usepackage{url}
\usepackage[hidelinks]{hyperref}
\usepackage[utf8]{inputenc}
\usepackage[small]{caption}
\usepackage{graphicx}
\usepackage{amsmath}
\usepackage{amsthm}
\usepackage{booktabs}
\usepackage{algorithm}
\usepackage{algorithmic}
\usepackage[switch]{lineno}
\usepackage{textcomp}
\usepackage{subfigure}
\usepackage{url}
\usepackage{verbatim}
\usepackage{graphicx}
\usepackage{amssymb}
\usepackage{amsmath}
\usepackage{array,booktabs}
\usepackage{multirow}
\usepackage{bm}
\usepackage{bbding}
\usepackage{pifont}
\usepackage{makecell}
\usepackage{xcolor}
\usepackage{array}
\usepackage{tabularx}
\usepackage{newfloat}
\usepackage{listings}
\usepackage{algorithm}
\usepackage{algorithmic}
\usepackage{hyperref}

\title{MFTraj: Map-Free, Behavior-Driven Trajectory Prediction for \\Autonomous Driving}

\author{
Haicheng Liao\textsuperscript{\rm 1}\thanks{Authors contributed equally; \dag Corresponding author.}\and
Zhenning Li\textsuperscript{\rm 1}$^{*\dag}$\and
Chengyue Wang\textsuperscript{\rm 1}\and
Huanming Shen\textsuperscript{\rm 2}\and
Bonan Wang\textsuperscript{\rm 1}\and\\
Dongping Liao\textsuperscript{\rm 1}\and
Guofa Li\textsuperscript{\rm 3}\and
Chengzhong Xu\textsuperscript{\rm 1}
\\
\affiliations
$^1$University of Macau\\
$^2$University of Electronic Science and Technology of China\\
$^3$Chongqing University\\
\emails
\{yc27979, zhenningli, chengyuewang, mc3500, yb97428, czxu\}@um.edu.com,
HuanmingShen@outlook.com.com,
hanshan198@gmail.com
}

\begin{document}

\maketitle

\begin{abstract}
This paper introduces a trajectory prediction model tailored for autonomous driving, focusing on capturing complex interactions in dynamic traffic scenarios without reliance on high-definition maps. The model, termed MFTraj, harnesses historical trajectory data combined with a novel dynamic geometric graph-based behavior-aware module. At its core, an adaptive structure-aware interactive graph convolutional network captures both positional and behavioral features of road users, preserving spatial-temporal intricacies. Enhanced by a linear attention mechanism, the model achieves computational efficiency and reduced parameter overhead. Evaluations on the Argoverse, NGSIM, HighD, and MoCAD datasets underscore MFTraj's robustness and adaptability, outperforming numerous benchmarks even in data-challenged scenarios without the need for additional information such as HD maps or vectorized maps. Importantly, it maintains competitive performance even in scenarios with substantial missing data, on par with most existing state-of-the-art models. The results and methodology suggest a significant advancement in autonomous driving trajectory prediction, paving the way for safer and efficient autonomous systems.
\end{abstract}

\section{Introduction}
The integration of autonomous vehicles (AVs) with human-driven vehicles and pedestrians necessitates advanced trajectory prediction models. Central to these models is their ability to predict the future trajectories of various road users, leveraging historical data. Despite significant advancements, a pivotal challenge persists: modeling the often unpredictable \textit{driving behaviors} of road users. 

These behaviors, shaped by a blend of traffic dynamics, road layouts, and individual cognitive inclinations, offer a unique window into the real-time decision-making processes of humans in complex traffic settings \cite{schwarting2019social,LiDCHCG23}. Our research has illuminated the pivotal role of understanding human behavioral patterns in trajectory predictions. Recognizing and predicting human driving behavior is not merely about tracing a vehicle's path; it's about understanding the cognitive processes that dictate those paths. By understanding behaviors, AVs can anticipate sudden changes in human-driven vehicles or pedestrian movements, leading to safer co-navigation. Furthermore, behavior-focused predictions can aid in scenarios where traditional data might be ambiguous or incomplete, relying on human behavioral patterns to fill in the gaps. Through the integration of decision-making theories, cognitive psychology, and traffic behavior studies \cite{yin2021multimodal}, trajectory prediction models can be enriched, fostering a harmonious coexistence of AVs and human-driven entities on the road.

High Definition (HD) maps, conventionally considered pivotal for trajectory prediction, pose intrinsic challenges. Their creation is resource-intensive, and in the rapidly changing milieu of urban environments, they can quickly become obsolete \cite{gao2020vectornet,ren2024emsin}. This has given rise to \textit{map-free} models, a paradigm shift that operates independently of HD map data. However, while these models adeptly handle dynamic environments, they may lack the granularity provided by comprehensive road network data. This gap is aptly addressed by the advent of deep learning techniques, notably Graph Neural Networks (GNNs) \cite{liang2020learning,gao2020vectornet}. GNNs, adept at assimilating extensive data from road users, offer nuanced insights into their interactions and the overarching socio-cognitive context, thereby compensating for the lack of detailed HD maps.

Our contributions are as follows:

\begin{enumerate}
    \item An \textbf{advanced map-free architecture} for trajectory prediction that obviates the need for HD maps, resulting in significant computational savings.
    \item  A novel dynamic geometric graph that captures the \textbf{essence of continuous driving behavior}, circumventing the limitations of manual labeling. We have integrated metrics and behavioral criteria, drawing from traffic psychology, cognitive neuroscience, and decision-making frameworks, to craft a model that offers more than mere predictions—it elucidates.
    \item  Benchmark assessments underscore MFTraj's superiority. Demonstrating a commendable \textbf{performance elevation of nearly 5.0\%} on the Argoverse, NGSIM, HighD, and MoCAD datasets, its robustness is further accentuated by its consistent performance even with a data shortfall of 25\%- 62.5\%, underscoring its adaptability and profound understanding of diverse traffic dynamics.
\end{enumerate} 
\section{Related Work}\label{Related work}
Recent years have seen an explosion of research in trajectory prediction for autonomous driving (AD), thanks to the burgeoning field of deep learning. These cutting-edge approaches \cite{10468619,messaoud2021attention,tian2024hydralora,liao2024physics} have demonstrated superior performance in complex traffic scenarios. However, they often encounter challenges in adequately representing spatial relationships such as graphic inputs of the scene. To address this, HD maps, rich in scene and semantic information, have attracted increasing research attention. Considering that Convolutional Neural Networks (CNNs) excel at extracting spatial features from input data, such as spatial features from inputs like vectorized maps or raster images, several studies \cite{zhao2021tnt,gilles2021home,khandelwal2020if} have merged sequential networks with CNNs. This hybrid approach effectively captures both temporal and spatial features from HD maps, providing enriched contextual information for motion prediction tasks. Recent research has explored Graph Neural Networks (GNNs) \cite{liang2020learning,zeng2021lanercnn,mohamed2020social,liao2024cognitive}, such as Graph Convolutional Networks (GCNs) and Graph Attention Networks (GATs), Transformers \cite{Zhou_2022_CVPR,liao2024human,chen2022vehicle,ngiam2022scene}, and generative models such as Adversarial Networks (GANs) \cite{zhou2023query} and Variational Auto Encoders (VAEs) \cite{walters2021trajectory,liao2024gpt} for direct encoding of HD maps. For example, VectorNet \cite{gao2020vectornet} simplifies maps by extracting key points from lane splines and encoding them using GNNs. Moreover, LaneGCN \cite{liang2020learning} and TPCN \cite{ye2021tpcn} build lane graphs using central line segments, employing GCN to capture dynamic interaction.  In addition, HiVT \cite{Zhou_2022_CVPR} and SSL-Lanes \cite{bhattacharyya2023ssl} represent map elements by relative positions, improving the transformer model for time series trajectory data. 

Despite their effectiveness, the limited availability of HD maps and the extensive resources needed for their creation and maintenance impede the widespread adoption of autonomous driving, particularly in areas lacking current HD map coverage. In response to these challenges, this study introduces a map-free model that utilizes generative models and a VRNN \cite{chung2015recurrent} to account for the variable nature of traffic scenarios. We propose a novel adaptive GCN to model the complexity of real-time interactions in traffic scenes. To streamline model complexity, we apply the Linformer framework \cite{wang2020linformer} for a balance between computational efficiency and prediction accuracy in AD applications.

\section{Methodologies}\label{Problem Formulation}
\subsection{Inputs and Outputs}
This study focuses on predicting the trajectory of the \textit{target vehicle} in interactive driving scenarios, considering all vehicles within the AV's (termed the \textit{target vehicle}) sensing range. At time \( t \), the ego vehicle anticipates the target vehicle's trajectory for the upcoming \( t_{f} \) steps. Our model, drawing from historical data \( \bm{X} \), considers past trajectories of both the target vehicle (indexed by \( 0 \)) and its surrounding agents (indexed from \( 1 \) to \( n \)) over a predefined horizon \( t_{h} \). Formally,

\begin{equation}\label{eq.1}
    \bm{X} = \left\{\bm{X}_{0}^{t-t_{h}:t}; \bm{X}_{i}^{t-t_{h}:t}\ \, \forall i\in[1,n] \right\}
\end{equation}
where $\bm{X}_{0}^{t-t_{h}:t}=\{x_{0}^{t-t_h},x_{0}^{t-t_h+1},\ldots, x_{0}^{t}\}$ and $\bm{X}_{i}^{t-t_{h}:t}=\{x_{i}^{t-t_h},x_{i}^{t-t_h+1},\ldots, x_{i}^{t}\}$ represent the historical trajectories of the target vehicle and those of the surrounding agents from time $t-t_{h}$ to $t$, respectively. 

The output of the model is the future trajectory of the target vehicle during the prediction horizon $t_{f}$:
\begin{equation}\label{eq.4}
    \bm{Y} = \left\{y_{0}^{t+1},y_{0}^{t+2},\ldots,y_{0}^{t+t_{f}-1},y_{0}^{t+t_{f}}\right\}
\end{equation}
where $y_{0}^{t}$ is the 2D coordinates of the target vehicle at time $t$.

\begin{figure*}[t]
  \centering
\includegraphics[width=0.75\textwidth]{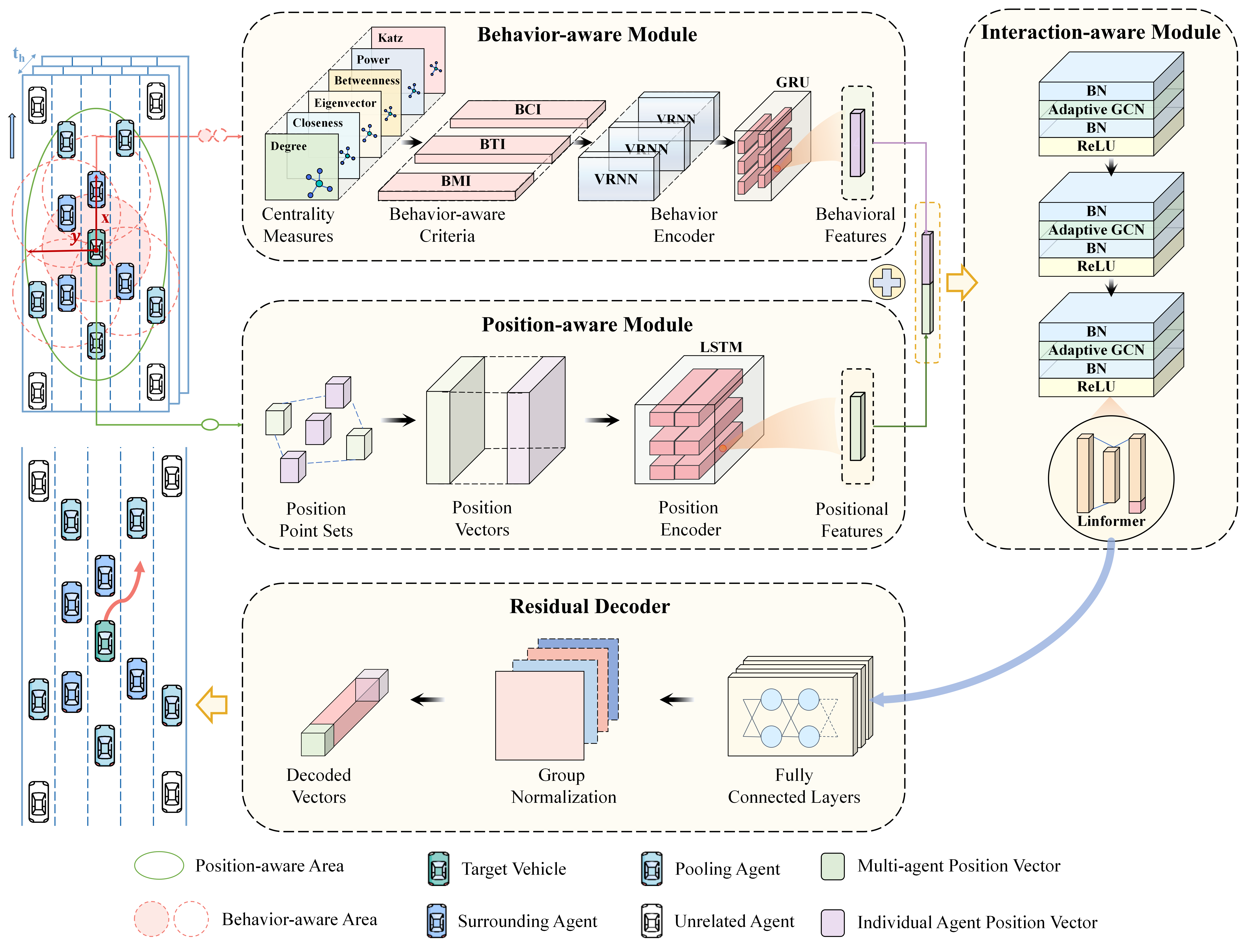} 
  \caption{Architecture of the proposed trajectory prediction model.}
  \label{fig1} 
\end{figure*}

Our model uniquely operates without relying on maps, using only the historical trajectory data of the target and nearby agents. The model needs an input sequence of length \( t_{h} \) and remains functional even if the historical data is not perfectly sequential. For sporadic missing data points, due to reasons like occlusions or sensor glitches, we employ simple linear imputation or similar methods. In addition, Figure \ref{fig1} illustrates our proposed model's hierarchical design. Following the encoder-decoder format, it features four key components: behavior-aware, position-aware, interaction-aware modules, and the residual decoder. We delve into each module's specifics below.

\subsection{Behavior-aware Module}\label{Behavior-aware Module_0}
Moving away from traditional methods that classify driver behaviors into fixed and discrete categories, we offer a more adaptable and flexible solution with our behavior-aware module, which utilizes a continuous portrayal of behavioral attributes. This approach draws inspiration from the multi-policy decision-making framework \cite{markkula2020defining}, integrating elements of traffic psychology \cite{toghi2022social} and dynamic geometric graphs (DGGs) \cite{boguna2021network} to effectively capture intricate driving behaviors amid ongoing driving maneuvers and evolving traffic conditions.\\ 
\textbf{Dynamic Geometric Graphs.} We first model the interactions of different agents with a DGG. At time $t$, the graph $G^{t}$ is defined as:
\begin{equation}\label{eq.7-1}
{G}^{t} = \{V^{t},{E}^{t}\}
\end{equation}
where $V^{t}=\{{v}_{0}^{t},{v}_{1}^{t}\ldots,{v}_{n}^t\}$ is the set of nodes, ${v}_{i}^{t}$ is the $i$-th node representing the $i$-th agent, ${E^{t}} = \{{e_{0}^{t}},{e_{1}^{t}}\ldots,{e_{n}^{t}}\}$ is the set of edges representing potential interactions between agents, and ${e_{i}^{t}}$ is the edge between the node ${v}_{i}^{t}$ and other agents that have potential influences with it. An interaction is assumed to exist only if two agents, e.g., $v_{i}$ and $v_{j}$, are in close proximity to each other, i.e., their shortest distance $d\left(v_{i}^{t}, v_{j}^{t}\right)$ is less than or equal to a predefined threshold $r$. Therefore, we define 
\begin{equation}\label{eq.7-4}
{e_{i}^{t}}= \{{v_{i}^{t}} {v_{j}^{t}} \mid(j \in  {N}_{i}^{t})\}
\end{equation}
where $ {N}_{i}^{t}=\left\{v_{j}^{t} \in V^{t}\backslash\left\{v_{i}^{t}\right\}\mid d\left( v_{i}^{t}, v_{j}^{t}\right) \leq r, i \neq j\right\}$.

Correspondingly, the symmetrical adjacency matrix $A^{t}$ of $G^{t}$ can be given as:
\begin{equation}\label{eq.8}
A^{t}(i, j)= \begin{cases}d\left(v_{i}^{t}, v_{j}^{t}\right) & \text { if } d\left(v_{i}^{t}, v_{j}^{t}\right)\leq{r}, i \neq j \\ 0 & \text {otherwise}\end{cases}
\end{equation}
\begin{table*}[htbp]
  \centering
  \caption{Centrality measures and their interpretations.}
  \resizebox{\linewidth}{!}{ 
    \begin{tabular}{cccc}
    \toprule
    Centrality Measures & \multicolumn{1}{c}{ Magnitude (Original Measure)} & Gradient (1st Derivative) & \multicolumn{1}{c}{Curvature (2nd Derivative)} \\
    \midrule
    \makecell{Degree $(\mathcal{J}_{i}^{t}(D))$\\Closeness $(\mathcal{J}_{i}^{t}(C))$} & \makecell{Agent's potential and capability \\for interaction in the traffic environment} & \makecell{Agent's sensitivity to \\traffic density variations} & \makecell{Driver's capability to react \\to fluctuating traffic conditions} \\
    \midrule
    \makecell{Eigenvector $(\mathcal{J}_{i}^{t}(E))$ \\Betweenness $(\mathcal{J}_{i}^{t}(B))$} & \makecell{Agent's significance \\in dynamic traffic scenarios} & \makecell{Variation in agent's importance \\in dynamic traffic scenes} & \makecell{Influence of driver behavior alterations \\on overall traffic conditions} \\
    \midrule
    \makecell{ Power $(\mathcal{J}_{i}^{t}(P))$\\Katz $(\mathcal{J}_{i}^{t}(K))$} & \makecell{Extent of influence an agent exerts \\on others via direct and indirect interactions at time $t$} & \makecell{Agent's adaptability to \\shifts in driving behaviors} & \makecell{Agent's capacity to modify interactions \\in complex and congested traffic scenarios} \\
    \bottomrule
    \end{tabular}%
    }
  \label{table}%
\end{table*}%
\textbf{Centrality Measures.}\label{Centrality Measures} Centrality measures are graph-theoretic metrics that are widely used to understand various aspects of network structures. These measures provide valuable insights into the importance, influence, and connectivity of nodes or vertices within a graph. As shown in Table 1, we use six centralities to characterize driving behavior. 
These measures allow evaluation of individual agent behavior within the DGG and reveal key agents and overall connectivity of the traffic graph.\\ 
\textit{(1) Degree Centrality:} Reflects the number of connections an agent has, correlating with its influence on and susceptibility to others. It's defined as:
\begin{equation}
    {J}_{i}^{t}(D)= \left|  {N}_{i}^{t}\right|+ {J}_{i}^{t-1}(D)
\end{equation}
where $\left| \mathcal{N}{i}^{t}\right|$ denotes the total elements in ${N}_{i}^{t}$. \\
\textit{(2) Closeness Centrality:} Indicates an agent's reachability, suggesting its potential influence over others. Defined by:
\begin{equation}
    {J}_{i}^{t}(C)=\frac{\left|  {N}_{i}^{t}\right|-1}{\sum_{\forall v_{j}^{t} \in  {N}^{t}_{i}}d\left(v_{i}^{t}, v_{j}^{t}\right)}
\end{equation}\\
\textit{(3) Eigenvector Centrality:} Measures an agent's importance by considering both quantity and quality of connections. Expressed as:
\begin{equation}
    {J}_{i}^{t}(E)=\frac{ \sum_{\forall v_{j}^{t} \in  {N}^{t}_{i}}d\left(v_{i}^{t}, v_{j}^{t}\right)}{\lambda}
\end{equation}
where $\lambda$ is the eigenvalue, indicating the collective influence exerted by an agent and its network.\\
\textit{(4) Betweenness Centrality:} Highlights agents that act as bridges or bottlenecks in traffic, crucial in congested situations. Formulated as:
\begin{equation}
    {J}_{i}^{t}(B) = \sum_{\forall v_{s}^{t},v_{k}^{t} \in {V}^{t}} \frac{\sigma_{j,k}(v_{i}^{t})}{\sigma_{j,k}}
\end{equation}
where $V^{t}$ denotes the set of all agents present in the scene, $\sigma_{j,k}$ signifies the total number of shortest paths between agent $v_{j}^{t}$ and agent $v_{k}^{t}$, and $\sigma_{j,k}(v_{i})$ represents the number of those paths traversing the agent $v_{i}^{t}$. \\
\textit{(5) Power Centrality:} Identifies agents in recurrent interactions, hinting at traffic patterns. Defined by:
\begin{equation}
    {J}_{i}^{t}(P) = \sum_{k}\frac{A^{k}_{ii}}{k!}
\end{equation}
where $A^{k}_{ii}$ denotes the $i$-th diagonal element of the adjacency matrix raised to the $k$-th power, while $k!$ signifies the factorial of $k$, shedding light on its contribution to the network's structural integrity and dynamism.\\
\textit{(6) Katz Centrality:} Emphasizes both direct and distant interactions of an agent, capturing intricate driving patterns. Given as:
\begin{small}
\begin{equation}
    {J}_{i}^{t}(K) = \sum_{k} \sum_{j} \alpha^{k} A^{k}_{ij}+\beta^{k},  \forall i, j \in[0,n], \text { where } \alpha^{k} <\frac{1}{\lambda_{\max }}
\end{equation}
\end{small}
where $n$ represents the number of agents in the real-time traffic scenario, $\alpha^{k}$ is the decay factor, $\beta^{k}$ denotes the weight assigned to the immediate neighboring agents, and $A^{k}_{ij}$ is the $i$,$j$-th element of the $k$-th power of the adjacency matrix.\\
\textbf{Behavior-aware Criteria.}
Inspired by the relationship between velocity, acceleration, and jerk, we introduce behavioral criteria. These criteria, consisting of Behavior Magnitude Index (BMI) ${\mathcal{C}}_{i}^{t}$, Behavior Tendency Index (BTI) $ {\mathcal{L}}_{i}^{t}$, and Behavior Curvature Index (BCI) $ {{\mathcal{I}}}_{i}^{t}$, evaluate different driving behaviors for the target vehicle and its surroundings. They compute thresholds, gradients, and concavities of centrality measures that reflect behaviors such as lane changes, acceleration, deceleration, and driving style. We find that behaviors with significant, fluctuating centrality values in short time frames are likely to affect nearby agents, consistent with human risk perception and time-sensitive decision-making.
They are respectively given as follows:
\begin{scriptsize}
\begin{equation}
{\mathcal{C}}_{i}^{t}=\left[\left|{ \mathcal{J}^{t}_{i}(D)}\right|,\left|{ \mathcal{J}^{t}_{i}(C)}\right|,\left|{ \mathcal{J}^{t}_{i}(E)}
\right|,
\left|{ \mathcal{J}_{i}^{t}(B)}\right|, \left|{ \mathcal{J}_{i}^{t}(P)}\right|, \left|{ \mathcal{J}_{i}^{t}(K)}\right|\right]^{T}
\end{equation}
\end{scriptsize}
{\begin{scriptsize}
\begin{flalign}\label{eq.14_1}
 {\mathcal{L}}_{i}^{t}=\left[\left|\frac{\partial \mathcal{J}^{t}_{i}(D)}{\partial t}\right|,\left|\frac{\partial \mathcal{J}^{t}_{i}(C)}{\partial t }\right|,
\cdots, \left|\frac{\partial \mathcal{J}_{i}^{t}(K)}{\partial t }\right|\right]^{T}
\end{flalign}
\end{scriptsize}}
\begin{scriptsize}
\begin{flalign}\label{eq.15}
    {{\mathcal{I}}}_{i}^{t} = \left[\left|\frac{\partial^{2} \mathcal{J}_{i}^{t}(D)}{\partial^{2} t}\right|,\left|\frac{\partial^{2} \mathcal{J}_{i}^{t}(C)}{\partial^{2} t }\right|,\cdots, \left|\frac{\partial \mathcal{J}_{i}^{t}(K)}{\partial^{2} t }\right|\right]^{T} 
\end{flalign}
\end{scriptsize}\\
\textbf{Behavior Encoder.}
Incorporating behavior-aware criteria, symbolized as \( \mathcal{J}=\{\mathcal{C}_{0:n}^{t-t_{h}:t},\mathcal{L}_{0:n}^{t-t_{h}:t},\mathcal{I}_{0:n}^{t-t_{h}:t}\} \), our behavior encoder comprises VRNN and GRU components. This encoder succinctly models relationships between random variables across time, yielding precise sequential behavioral features \( {\bar{O}_{\textit{behavior}}}^{t-t_{h}:t} \). Formally:
\begin{equation}
 {\bar{O}_{\textit{behavior}}}^{t-t_{h}:t} = \phi_{ \textit{GRU}} \left( \phi_{\textit{VRNN}}(\mathcal{J}) \right)
\end{equation}
where \( \phi_{ \textit{GRU}} \) and \( \phi_{ \textit{VRNN}} \) denote the GRU and VRNN functions. This encoder captures human driving patterns and their temporal dynamics. Next, behavioral features \( {\bar{O}_{\textit{behavior}}}^{t-t_{h}:t} \), fuse with positional features from the position-aware module, subsequently processed by the interaction-aware module for comprehensive feature extraction. 

\subsection{Position-aware Module}\label{Position-aware Pooling Module_0}
Contrary to traditional methods that emphasize absolute positions \cite{wang2022ltp,gao2020vectornet} or fixed grids \cite{deo2018convolutional}, our model emphasizes relative positions. The position-aware module captures individual and group spatial dynamics, interpreting the scene's geometric nuances. These insights are then encoded to produce positional features.
\begin{figure*}[t]
  \centering
\includegraphics[width=0.9\linewidth]{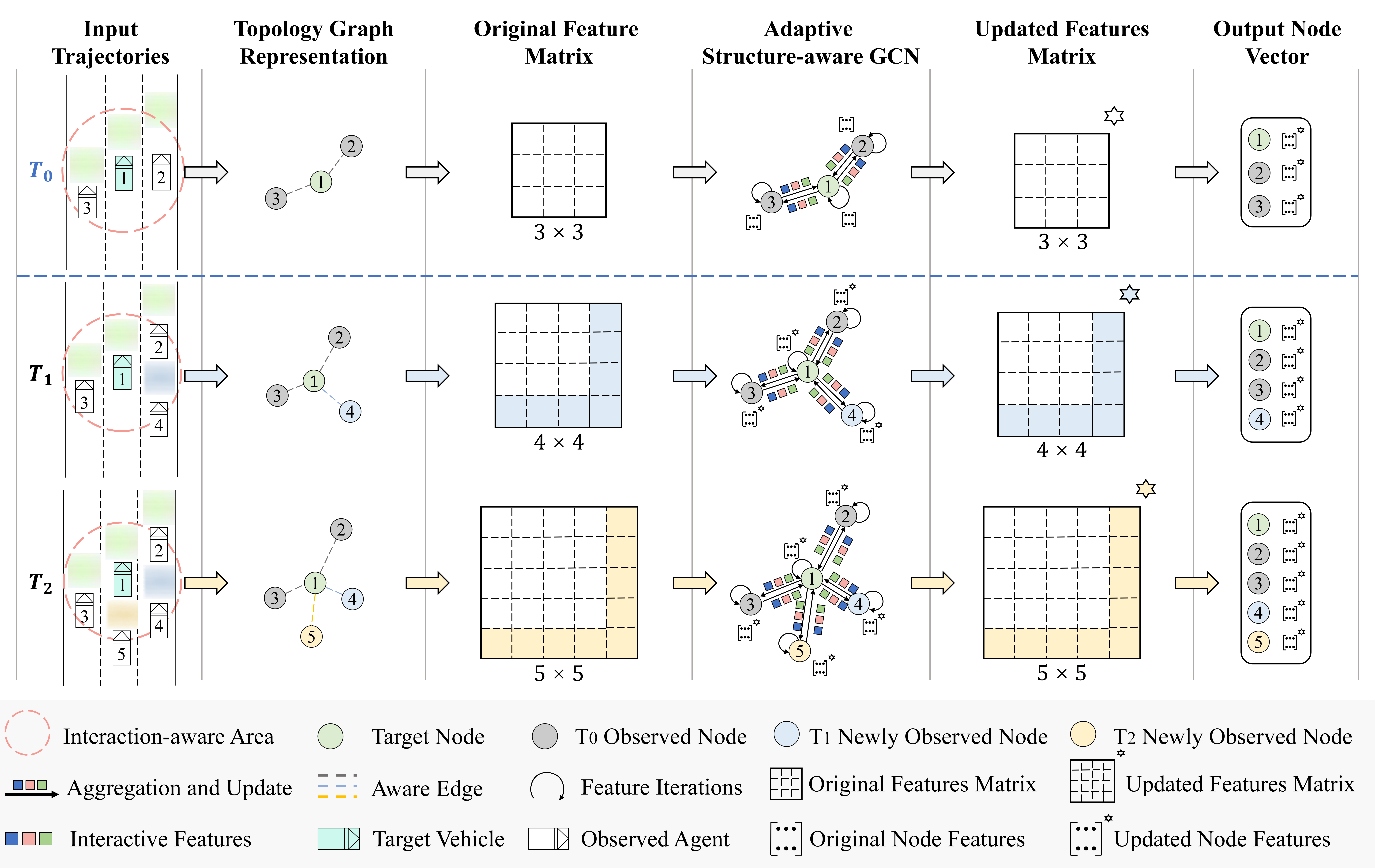} 
  \caption{Overview of our adaptive structure-aware GCN. The real-time trajectories of the target and observed agents are captured using a topology graph to form a feature matrix. This matrix undergoes aggregation, updating, and iteration within the GCN. As new agents are observed in real-time, the GCN dynamically adjusts its topology, updating features for the added nodes.}
  \label{fig2} 
\end{figure*}\\
\textbf{Pooling Mechanism.} Our pooling mechanism captures dynamic position data from the traffic environment around the target vehicle, utilizing individual \( \mathbf{s}_{i}^{t_{k}} \) and multi-agent \( \mathbf{s}_{i, j}^{t_{k}} \) position vectors. This strategy gleans historical trajectories and spatial relationships without depending on fixed positions or grids. The relationships are formulated as:
\begin{equation}
\mathbf{s}_{i}^{t_{k}} = \{ p_{i}^{t_{k}} - p_{i}^{t_{k}-1} \}, \, \, \mathbf{p}_{i, j}^{t_{k}} = \{ p_{i}^{t_{k}} - p_{j}^{t_{k}} \}
\end{equation}\\
\textbf{Position Encoder.}
The position encoder employs an LSTM to transform discrete position vectors into continuous spatio-temporal representations, thereby enhancing temporal and spatial interactions between agents and map elements. Given the historical position vectors for the target and surrounding agents, it embeds them temporally:
\begin{equation}
 {\bar{O}}_{\textit{position}}^{t-t_{h}:t}=\phi_{ \textit{LSTM}}\left(\mathbf{\bar{h}}^{t-t_{h}:t-1}_{i}, \mathbf{s}^{t-t_{h}:t-1}_{i}, \mathbf{{p}}^{t-t_{h}:t-1}_{i,j}\right)
\end{equation}
where $ {\bar{O}}_{\textit{position}}^{t-t_{h}: t-1}$ is the positional features output by the position encoder, and 
 $\phi_{\textit{LSTM}}$ denotes the two-layer LSTM encoder, and $\mathbf{\bar{h}}^{t-t_{h}:t-1}_{i}$ represents the hidden position state updated by the encoder on a frame-by-frame basis, with the weights of the LSTM shared among all agents.

\subsection{Interaction-aware Module}\label{Interaction-aware Pooling Module}
Effective trajectory prediction in complex traffic scenarios hinges upon a system's ability to comprehend and anticipate interactions among vehicles. Classic GCN-based methods, although proficient at encapsulating geometric inter-agent relationships, often exhibit limitations in fluid traffic conditions due to their fixed adjacency matrix configurations. To tackle this, we introduce a novel adaptive structure-aware GCN, taking cues from advancements in crystal graphs and material design. This novel approach stands out by its capability to craft spatial feature matrices dynamically, adjusting to the number of agents observed in real-time, which ensures a more fluid and adaptable response to traffic changes. A graphical illustration of this concept is provided in Figure \ref{fig2}.

Breaking away from conventional models that predominantly lean on distance-based positional features, our design holistically blends continuous behavioral features into its graph structure. This not only addresses the multifaceted spatio-temporal interactions but also considers the intricate physical interplays between agents, offering a noticeable enhancement in prediction precision. Our design blueprint encompasses an adaptive convolutional neural network rooted in a fully connected interaction multigraph. This structure is adept at simultaneously capturing sequential behavioral and dynamic positional interactions among agents. The multigraph's operational layer is distinguished by nodes, which symbolize sequential behavioral features $ {\bar{O}}_{\textit{behavior}}^{t-t{h}:t}$ and edges representing positional features $ {\bar{O}}_{\textit{position}}^{t-t{h}:t}$, as defined below:
\begin{equation}
\begin{scriptsize}
\begin{aligned}
\mathbf{\tilde{z}}_{i}^{k} & =  {F}\left(\mathbf{\tilde{z}}_{i}^{k-1}, \tilde{r}_{i,j}^{k-1}\right) \\
& ={\mathbf{\tilde{z}}}_{i}^{k-1} +\phi_{\text {sgm}}\left(\mathbf{\tilde{r}}_{i,j} ^{k-1}\mathbf{W}_{\mathrm{g}}^{k-1}+\mathbf{\mathbf{b}}_{\mathrm{g}}^{k-1}\right) \odot \phi_{\text {spu}}\left(\mathbf{\tilde{r}}_{i,j} ^{k-1}\mathbf{W}_{\mathrm{h}}^{k-1}+\mathbf{\mathbf{b}}_{\mathrm{h}}^{k-1}\right) \\
\end{aligned}
\end{scriptsize}
\end{equation}
where the variable $k$ denotes the layer within the GCN, $k \in [1,3]$, and the symbols $\odot$, $\phi_{\text {sgm}}$, and $\phi_{\text {spu}}$ represent the element-wise product, sigmoid activation function, and softplus activation function, respectively.  Consequently, $\mathbf{W}_{\mathrm{g}}^{k-1}$ and $\mathbf{W}_{\mathrm{h}}^{k-1}$ are learnable matrices,
$\mathbf{\mathbf{b}}_{\mathrm{g}}^{k-1}$, and $\mathbf{\mathbf{b}}_{\mathrm{h}}^{k-1}$ are the bias of the $k-$th layer. $\tilde{r}_{i,j}^{k-1}$ can be represented as follows:
\begin{equation} 
\tilde{r}_{i,j}^{k-1} =\left(\mathbf{\tilde{z}}_{i}^{k-1}\left\|\mathbf{\tilde{z}}_{j}^{k-1}\right\|\mathbf{\mathbf{p}}_{i, j}^{t-t_{h}:t}\right)
\end{equation}
Additionally, the initial feature vector $\mathbf{\tilde{z}}_{i}^{(0)}$ is defined as follows:
\begin{equation} 
\mathbf{\tilde{z}}_{i}^{(0)} =\left(  {\bar{O}_{\textit{behavior}}}^{t-t_{h}:t}\| {\bar{O}}_{\textit{position}}^{t-t_{h}:t}\right)
\end{equation}
Furthermore, the output of the adaptive structure-aware GCN for the target vehicle $i$ is then passed to Linformer, an extension architecture of Transformer,

Furthermore, the output of the adaptive structure-aware GCN for the target vehicle $i$ is subsequently fed into a lightweight transformer-based framework—— Linformer \cite{wang2020linformer}, to efficiently quantify and compute the dynamic attention weight vectors for the surrounding agents, ultimately output the contextual mapping $ {\bar{O}}$.
This allows for a favorable trade-off between accuracy and efficiency.

\subsection{Residual Decoder}
The residual decoder, comprising a linear residual and projection layer, processes node vectors to forecast the target vehicle's future trajectory, producing the prediction $\mathbf{Y}_{0}^{t:t+t{f}}$. This is given by:
\begin{equation}
\bm{Y}=\boldsymbol{\mathbf{Y}_{0}^{t:t+t_{f}}}= {F}_{\theta}\left( {F}_{\theta}( {\bar{O}})\right)
\end{equation}
such that,
\begin{equation}
  {F}_{\theta}(\mathbf{\cdot})=\phi_{\text{ReLU}}\left[\phi_{\text {GN}}\left(\phi_{\text {Linear}}(\mathbf{\cdot})\right)\right]
\end{equation}
where $\phi_{\text {ReLU}}$ denotes the ReLU activation function, and $\phi_{\text{GN}}$ denotes the Group Normalization (GN) function \cite{wu2018group}, which is applied to improve the training stability of our model. 
In addition, the $\phi_{\text {Linear}}$ corresponds to the fully connected layer, while $ {F}_{\theta}$ denotes the residual decoder function.

\begin{table}[htbp]
  \centering
  \caption{Performance comparison of various models on \textit{complete} and \textit{missing} datasets for Argoverse. Models use either HD map or vectorized map (Map) and trajectory (Traj.) data or solely Trajectory data, with some not specifying ('-'). Metrics include minADE (k=1), minFDE (k=1), and MR (k=1). \textbf{Bold} and \underline{underlined} values represent the best and second-best performance in each category.}
   \resizebox{\linewidth}{!}{
    \begin{tabular}{ccccc}
    \toprule
   Model & \multicolumn{1}{c}{Input} & minADE (m)↓ & minFDE (m)↓ & MR (\%)↓ \\
    \midrule
    Argoverse Baseline \cite{chang2019argoverse} & Map + Traj. & \makecell{2.96}  & \makecell{6.81}  & \makecell{81.00}   \\
     Constant Velocity \cite{chang2019argoverse}& -     & \makecell{3.55}  & \makecell{7.89}  & \makecell{75.00}   \\
    SGAN \cite{gupta2018social} & - & 3.61  & 5.39  & 87.11   \\
    TPNet \cite{fang2020tpnet}& Map + Traj. & 2.33  & 5.29  & -     \\
    PRIME  \cite{song2022learning}& Map + Traj. & 1.91  & 3.82  & 58.67 \\
    Uulm-mrm (2rd) \cite{chang2019argoverse}& Map + Traj. & 1.90  & 4.19  & 63.47   \\
    Jean (1st) \cite{mercat2020multi} & Map + Traj. & 1.74  & 4.24  & 68.56   \\
     WIMP \cite{khandelwal2020if} & Map + Traj. & 1.82  & 4.03  & 62.88    \\
    Scene-Transformer \cite{ngiam2022scene} & Map + Traj. & 1.81  & 4.06  & 59.21    \\
    TNT \cite{zhao2021tnt} & Map + Traj. & 1.77  & 3.91  & 59.72  \\
    mmTransformer \cite{liu2021multimodal}& Map + Traj. & 1.77  & 4.00  & 61.78   \\
    CtsConv (Aug.) \cite{walters2021trajectory} & Map + Traj. & 1.77  & 4.05  & -   \\
    HOME \cite{gilles2021home} & Map + Traj. & 1.72  & 3.73  & 58.40  \\
    LaneGCN  \cite{liang2020learning} & Map + Traj. & 1.71  & 3.78  & 59.05   \\
    GOHOME \cite{gilles2022gohome}& Map + Traj. & 1.69  & 3.65  & 57.21  \\
    LaneRCNN \cite{zeng2021lanercnn} & Map + Traj. & 1.68  & 3.69  & 56.85   \\
    DenseTNT \cite{gu2021DenseTNT} & Map + Traj. & 1.68  & 3.63  & 58.43  \\
    VectorNet \cite{gao2020vectornet}& Map + Traj. & 1.66  & 3.67  & -  \\
    TPCN \cite{ye2021tpcn}  & Map + Traj. & 1.66  & 3.69  & 58.80  \\
    SSL-Lanes \cite{bhattacharyya2023ssl} & Map + Traj. & 1.63  & 3.56  & 56.71   \\
    LTP \cite{wang2022ltp} & Map + Traj. & \multicolumn{1}{c}{1.62}     & \multicolumn{1}{c}{3.55}     & \underline{56.25}      \\
    {HiVT-128} \cite{Zhou_2022_CVPR} & Map + Traj. & \underline{1.60}     & \underline{3.52}     & -     \\
    \midrule
    \textbf{MFTraj} & Traj. & {\textbf{1.59}} & {\textbf{3.51}} &{\textbf{55.44}}   \\
    MFTraj  (drop 3-frames) & Traj. & 1.68  & 3.59  &56.95   \\
    MFTraj (drop 5-frames) & Traj. & 1.76  & 3.74  & 59.08  \\
    MFTraj (drop 8-frames) & Traj. & 1.86  & 3.90  & 61.12  \\
    MFTraj (drop 10-frames) & Traj. & 1.97  & 3.96  &62.72  \\
    \bottomrule
    \end{tabular}%
    }
  \label{Table-1}%
\end{table}%

\section{Experiments}\label{Experiments}
\subsection{Experimental Setup}
\textbf{Datasets.} We tested model's efficacy on Argoverse \cite{chang2019argoverse}, NGSIM \cite{deo2018convolutional}, HighD \cite{highDdataset}, and MoCAD \cite{liao2024bat} datasets.\\
\textbf{Data Segmentation.} For Argoverse, we predicted a 3-second trajectory from a 2-second observation, while for NGSIM, HighD, and MoCAD, we use 6-second intervals split into 2 seconds of observation and 4 seconds of prediction. These datasets, referred to as the \textit{complete} dataset, help assess our model in diverse traffic scenarios. Recognizing that real-world conditions often lead to incomplete data, we further assessed our model's resilience using the Argoverse dataset by introducing four subsets with varying levels of missing data: \textit{drop 3-frames}, \textit{drop 5-frames}, \textit{drop 8-frames}, and \textit{drop 10-frames}. These \textit{missing} datasets simulate data loss scenarios. For data gaps, we applied simple linear interpolation. \\
\textbf{Metrics.} Our experimental protocol was aligned
with the Argoverse Motion Forecasting Challenge and prior work \cite{liao2024bat}, we evaluated the performance of our model using standard metrics: minADE, minFDE, MR, and RMSE.  \\
\textbf{Implementation Details.} We implemented our model using PyTorch and PyTorch-lightning on an NVIDIA DGX-2 with eight V100 GPUs. Using the smooth L1 loss as our loss function, the model was trained with the Adam optimizer, a batch size of 32, and learning rates of \(10^{-3}\) and \(10^{-4}\). 

\subsection{Experimental Results}
\textbf{Performance Evaluation on the \textit{Complete} Dataset.}
Tables \ref{Table-1} and Table \ref{table_overall} present a comparative evaluation of our trajectory prediction model against 25 baselines from 2016 to 2023. Unlike most approaches that depend on HD maps or vectorized map data, our model omits map-based inputs. Still, it consistently outperforms the baselines across metrics like minADE, minFDE, MR, and RMSE for both Argoverse and MoCAD datasets. Specifically, for the Argoverse dataset, MFTraj outperforms most of the SOTA models by margins of 2.9\% in minADE, 2.4\% in minFDE, and 3.8\% in MR, while being on par with HiVT. It excels particularly in challenging long-term predictions (4s-5s) on NGSIM, HighD, and MoCAD datasets, with reductions in forecast error surpassing at least 11.5\%, 29.6\%, and 21.9\%, respectively. This emphasizes its potential for accurate long-term predictions in highway and urban settings. \\
\textbf{Performance Evaluation on the \textit{Missing} Dataset.}
Table \ref{Table-1} showcases the resilience of our model when faced with incomplete data sets. Our model consistently outperforms all other baselines on the \textit{drop 3-frames} and \textit{drop 5-frames} datasets. Notably, on the \textit{drop 3-frames} dataset, it surpasses nearly all state-of-the-art (SOTA) models trained on full data, highlighting its remarkable predictive strength even with missing data. While its performance on the \textit{drop 5-frames} dataset excels over most baselines, there are exceptions in specific metrics against models like TNT, WIMP, and mm Transformer. As the number of missing frames increases, as in the \textit{drop 8-frames} and \textit{drop 10-frames} datasets, there's an expected decline in performance. Yet, even with half the input data missing, our model still competes strongly against top baselines, emphasizing its potential in environments with data interruptions.\\
\textbf{Comparative Analysis of Model Performance and Complexity.}
In Table \ref{Table_3}, we compare our model's performance and complexity with various SOTA baselines. While our model doesn't have the lowest parameter count, it excels in all performance metrics. Impressively, it achieves this while using 90.42\% and 87.18\% fewer parameters than WIMP and Scene-Transformer, respectively. Compared to top-10 SOTA models, our model not only surpasses them in accuracy but is also as efficient, if not more so, than HiVT-128, SSL-Lanes, LaneGCN, and HOME+GOHOME. This underlines our model's optimal balance of robustness, efficiency, and trajectory prediction accuracy.

\begin{table}[htbp]
  \centering
  \caption{Comparative evaluation of MFTraj with SOTA baselines.}
   \resizebox{\linewidth}{!}{
    \begin{tabular}{ccccc}
    \toprule
    Model & minADE (m)↓ & minFDE (m)↓ & MR (\%)↓ & \#Param (K) \\
    \midrule
    WIMP \cite{khandelwal2020if} &1.82  &4.03  & 62.88  &\multicolumn{1}{c}{\textgreater20,000} \\
    Scene-Transformer \cite{ngiam2022scene} & 1.81  & 4.06  & 59.21  &\multicolumn{1}{c}{15,296}  \\
    CtsConv (Aug.) \cite{walters2021trajectory} & 1.77  & 4.05  & \multicolumn{1}{c}{-}     &\multicolumn{1}{c}{\underline{1,078}}  \\
    mmTransformer \cite{liu2021multimodal} & 1.77  & 4.00  & 61.78  &\multicolumn{1}{c}{2,607}  \\
    LaneGCN \cite{liang2020learning} & 1.71  & 3.78  & 59.05  &\multicolumn{1}{c}{3,701}  \\
    HOME+GOHOME \cite{gilles2022gohome} & 1.69  & 3.65  & 57.21  &\multicolumn{1}{c}{5,100}  \\
    DenseTNT \cite{gu2021DenseTNT} & 1.68  & 3.63  & 58.43  &\multicolumn{1}{c}{1,103}  \\
    SSL-Lanes \cite{bhattacharyya2023ssl} & 1.63  & 3.56  & 56.71  &\multicolumn{1}{c}{1,840} \\
    HiVT-128  \cite{Zhou_2022_CVPR} & 1.60  & 3.52  & -     &\multicolumn{1}{c}{2,529}  \\
    \midrule
    MFTraj   & \underline{\textbf{1.59}} & \underline{\textbf{3.51}} & \underline{\textbf{55.44}} & \multicolumn{1}{c}{\textbf{1,961}} \\
    \bottomrule
    \end{tabular}%
    }
      \label{Table_3}%
\end{table}%

\begin{table}[htbp]
  \centering
     \caption{Evaluation results for MFTraj and the other SOTA baselines without using HD maps in the NGSIM, HighD and MoCAD datasets over a different horizon. RMSE (m) is the evaluation metric, with some not specifying (``-''). \textbf{Bold} and \underline{underlined} values represent the best and second-best performance in each category.}\label{Table1}
     \setlength{\tabcolsep}{3mm}
   \resizebox{\linewidth}{!}{
      \begin{tabular}{c|cccccc}
    \toprule
    \multicolumn{1}{c}{\multirow{2}[2]{*}{Dataset}} & \multirow{2}[3]{*}{Model} & \multicolumn{5}{c}{Prediction Horizon (s)} \\
\cmidrule{3-7}    \multicolumn{1}{c}{} &       & 1     & 2     & 3     & 4     & 5 \\
      \hline
    \multirow{12}[13]{*}{NGSIM} & S-LSTM \cite{alahi2016social} & 0.65  & 1.31  & 2.16  & 3.25  & 4.55 \\
          & S-GAN \cite{gupta2018social} & 0.57  & 1.32  & 2.22  & 3.26  & 4.40 \\
          & CS-LSTM \cite{deo2018convolutional} & 0.61  & 1.27  & 2.09  & 3.10  & 4.37\\
          & DRBP\cite{gao2023dual} & 1.18  & 2.83  & 4.22  & 5.82  & -   \\
          & DN-IRL \cite{fernando2019neighbourhood} & 0.54  & 1.02  & 1.91  & 2.43  & 3.76\\
          & WSiP \cite{wang2023wsip} & 0.56  & 1.23  & 2.05  & 3.08  & 4.34  \\
          & CF-LSTM \cite{xie2021congestion} & 0.55  & 1.10  & 1.78  & 2.73  & 3.82 \\
          & MHA-LSTM \cite{messaoud2021attention} & 0.41  & 1.01  & 1.74  & 2.67  & 3.83\\
          & HMNet \cite{xue2021hierarchical} & 0.50  & 1.13  & 1.89  & 2.85  & 4.04  \\
          & TS-GAN \cite{wang2022multi} & 0.60  & 1.24  & 1.95  & 2.78  & 3.72 \\
          & Stdan \cite{chen2022intention} & 0.39  & 0.96  & 1.61  & 2.56 & 3.67 \\
          & iNATran \cite{chen2022vehicle} & \underline{0.39}  & \underline{0.96}  & \underline{1.61}  & \underline{2.42}  & 3.43  \\  
          & DACR-AMTP \cite{cong2023dacr}& 0.57  & 1.07  & 1.68  & 2.53  & \underline{3.40}   \\ 
          & FHIF \cite{zuo2023trajectory} &0.40  & 0.98  & 1.66  & 2.52  & 3.63 \\  
          & \textbf{MFTraj} & \textbf{ 0.38 } & \textbf{ 0.87 } & \textbf{ 1.52 } & \textbf{ 2.23 } & \textbf{ 2.95 } \\
      \hline
    \multirow{7}[23]{*}{HighD} 
    &S-GAN \cite{gupta2018social}& 0.30  & 0.78  & 1.46  & 2.34  & 3.41  \\
    &WSiP \cite{wang2023wsip}& 0.20  & 0.60  & 1.21  & 2.07  & 3.14  \\
    &CS-LSTM \cite{deo2018convolutional}& 0.22  & 0.61  & 1.24  & 2.10  & 3.27\\
    &MHA-LSTM \cite{messaoud2021attention}& 0.19  & 0.55  & 1.10  & 1.84  & 2.78\\
    &NLS-LSTM \cite{messaoud2019non}& 0.20  & 0.57  & 1.14  & 1.90  & 2.91 \\
    &DRBP\cite{gao2023dual}& 0.41  & 0.79  & 1.11  & 1.40  & - \\
    &EA-Net \cite{cai2021environment} & 0.15  & 0.26  & 0.43  & 0.78  & 1.32  \\
    &CF-LSTM \cite{xie2021congestion}& 0.18  & 0.42  & 1.07  & 1.72  & 2.44 \\
    &Stdan \cite{chen2022intention}& 0.19  & 0.27  & 0.48  & 0.91  & 1.66  \\
    &iNATran \cite{chen2022vehicle}& \textbf{0.04}  & \textbf{0.05}  & \underline{0.21}  & {0.54}  & 1.10 \\
     &DACR-AMTP \cite{cong2023dacr}& 0.10  & 0.17  & 0.31  & {0.54}  & {1.01}\\
     &GaVa \cite{liao2024human}& 0.17  & 0.24  & 0.42  & 0.86  & 1.31  \\ 
    & \textbf{MFTraj} & \underline{0.07}  & \underline{0.10}  & \textbf{ 0.19 } & \textbf{ 0.38 } & \textbf{ 0.56 }  \\
     \hline
    \multirow{7}[13]{*}{MoCAD}
    &S-GAN \cite{gupta2018social} & 1.69  & 2.25  & 3.30  & 3.89  & 4.69 \\
    &CS-LSTM \cite{deo2018convolutional} & 1.45  & 1.98  & 2.94  & 3.56  & 4.49 \\
    &MHA-LSTM \cite{messaoud2021attention} & 1.25  & 1.48  & 2.57  & 3.22  & 4.20  \\
    &NLS-LSTM \cite{messaoud2019non} & 0.96  & 1.27  & 2.08  & 2.86  & 3.93  \\
    &WSiP \cite{wang2023wsip} & 0.70  & 0.87  & 1.70  & 2.56  & 3.47 \\
    &CF-LSTM \cite{xie2021congestion} & 0.72  & 0.91  & 1.73  & 2.59  & 3.44 \\
    &Stdan \cite{chen2022intention} &{0.62}  & {0.85}  &{1.62}  &{2.51}  & {3.32} \\
    &HLTP \cite{10468619}  & {0.55} & {0.76} & {1.44} & {2.39} & {3.21} \\
    &BAT \cite{liao2024bat} & \underline{0.35}  & \underline{0.74}  & \underline{1.39}  & \underline{2.19} &\underline{2.88}\\
    & \textbf{MFTraj} & \textbf{0.34}  & \textbf{0.70}  & \textbf{1.32}  & \textbf{2.01}  & \textbf{2.57} \\
    \bottomrule
    \end{tabular}%
  \label{table_overall}%
  }
\end{table}%

\subsection{Ablation Studies} \label{Ablation Studies}
We executed an ablation study to assess the impact of individual components within our trajectory prediction model, with the results summarized in Table \ref{Table3}. Model F, i.e., MFTraj, which integrates all components, stands out in all metrics, signifying the synergy of its parts.
When the behavior-aware module is excluded in Model A, there are noticeable drops in minADE, minFDE, and MR by 12.6\%, 8.8\%, and 8.5\% respectively, highlighting its pivotal role. Model B, with absolute coordinates, underperforms, emphasizing the relevance of spatial relationships. Model C, without the interaction-aware module and Linformer extension, and Model D, lacking Linformer, both show diminished performance. Similarly, Model E, which uses a standard GCN instead of the adaptive one, also lags, underscoring the latter's efficiency. In essence, this study solidifies the importance of each component in Model F. Every part, from understanding behavioral nuances to updating features effectively, bolsters the model's precision and resilience. In essence, this study solidifies the importance of each component in Model F. Every part, from understanding behavioral nuances to updating features effectively, bolsters the model's precision and resilience.
\vspace{-8pt}
\begin{figure}[htbp]
  \centering
\includegraphics[width=\linewidth]{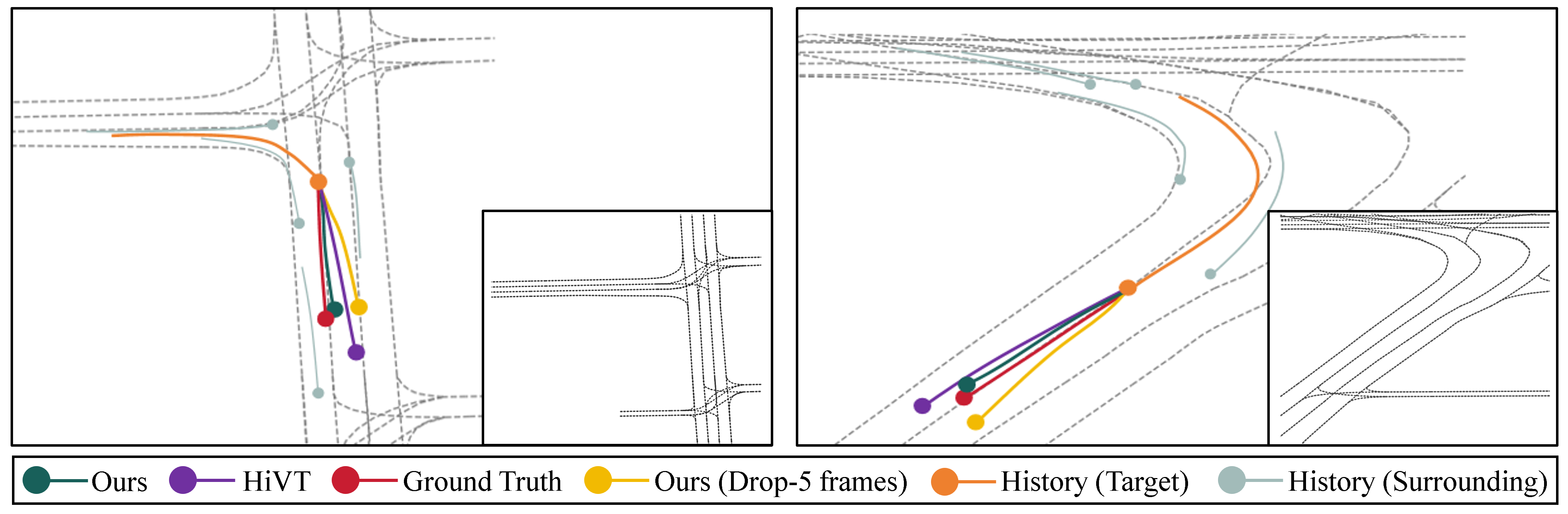} 
  \caption{Qualitative results of MFTraj and HiVT on Agroverse.}
  \label{fig2_1} 
\end{figure}

\begin{table}[htbp]
  \centering
  \caption{Ablation analysis of individual components in Argoverse.}
  \setlength{\tabcolsep}{5mm}
\resizebox{\linewidth}{!}{
    \begin{tabular}{cccc}
    \toprule
    \makecell{Ablation Models \\ $(\Delta{ Model \ F})$}  & \makecell{minADE (m)↓} & \makecell{minFDE (m)↓} & \makecell{MR (\%)↓}  \\
    \midrule
    \makecell{Model A} & \makecell{1.82} & \makecell{3.85} & \makecell{60.61}\\
    \makecell{Model B} &\makecell{1.69} &\makecell{3.59} &\makecell{56.14} \\
    \makecell{Model C} &\makecell{1.78} &\makecell{3.71} &\makecell{59.07} \\
    \makecell{Model D} &\makecell{1.71} &\makecell{3.61} &\makecell{57.59} \\
    \makecell{Model E} &\makecell{1.68} &\makecell{3.70} &\makecell{56.94} \\
    \midrule
    \makecell{Model F} & \makecell{\textbf{1.59}}  & \makecell{\textbf{3.51}}  & \makecell{\textbf{55.44}} \\
    \bottomrule
    \end{tabular}
    }
  \label{Table3}
\end{table}%

\begin{figure}[htbp]
  \centering
\includegraphics[width=\linewidth]{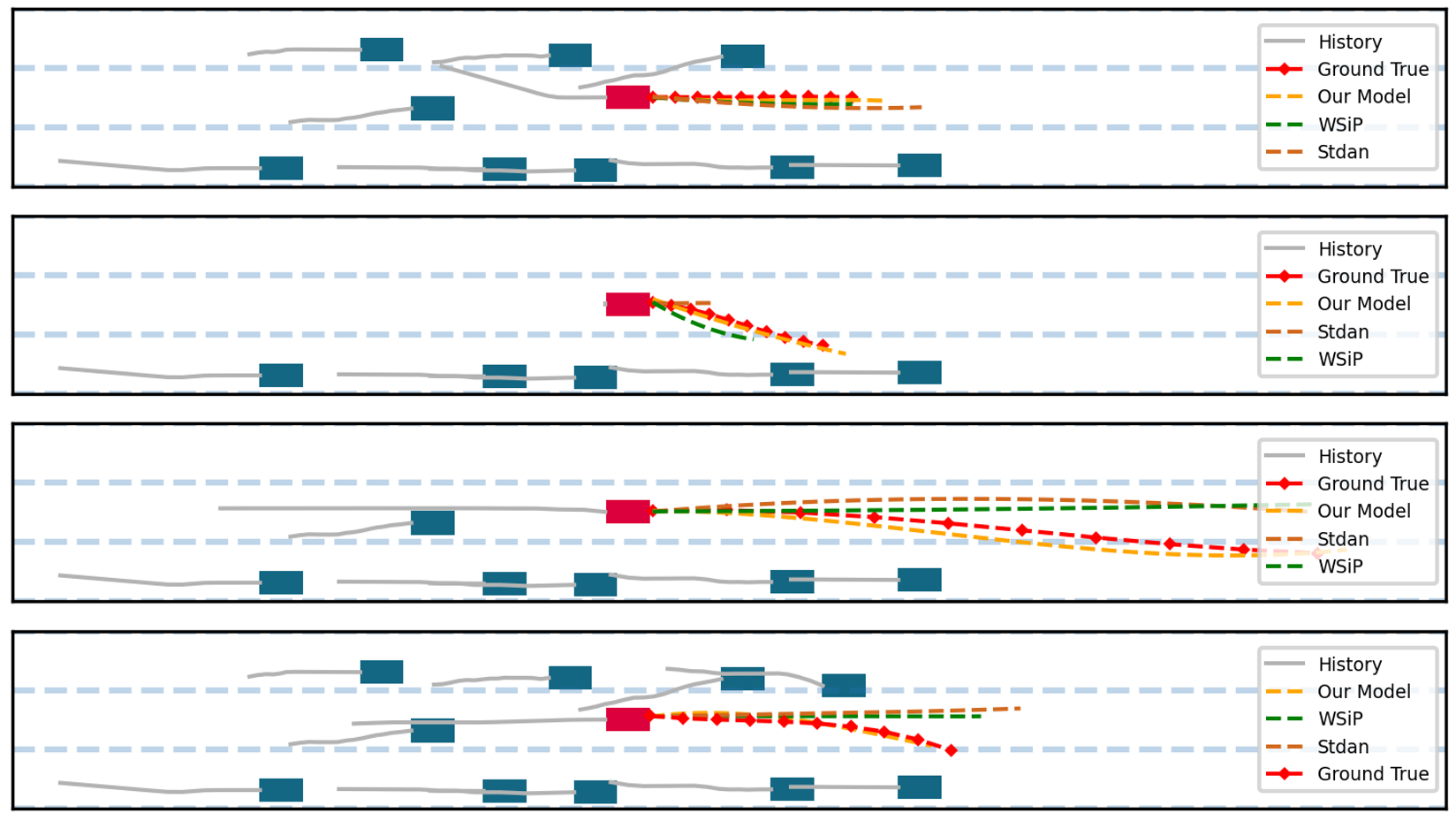} 
  \caption{Qualitative results of MFTraj on NGSIM. Target vehicle is depicted in red, while its surrounding agents are shown in blue.}
  \label{fig3} 
\end{figure}

\subsection{Qualitative Results}
Figure \ref{fig2_1} presents the qualitative results of our model using the Argoverse dataset. We've limited the display to the target vehicle's trajectories for clarity. Interestingly, without the aid of HD maps, our model adeptly discerns road semantics, enabling it to make precise and logical predictions for target vehicles in intricate urban settings. Importantly, Figure \ref{fig3} illustrates a comparison between the trajectories predicted by MFTraj and the SOTA baselines in the same traffic scenarios. MFTraj outperforms Stdan and WSiP in trajectory prediction, especially in complex scenarios such as lane changes and merging. These results demonstrate the superior adaptability and reliability of MFTraj in complex traffic conditions.

\section{Conclusion}\label{Conclusion}
This work presents a map-free and behavior-aware trajectory prediction model for AVs, integrating four components: behavior-aware, position-aware, interaction-aware modules, and a residual decoder. These components work in concert to analyze and interpret various inputs, understand human-machine interactions, and account for the inherent uncertainty and variability in the prediction. Evaluated with the Argoverse, NGSIM, HighD, and MoCAD datasets, MFTraj outperformed  SOTA baselines in prediction accuracy and efficiency without additional map information. Furthermore, this approach ensures its robustness and adaptability even in the presence of significant missing data; it achieved impressive performance even with a 50\% sequential input data deficit. This underscores the resilience and efficiency of MFTraj in predicting future vehicle trajectories and suggests its potential to drastically reduce the data requirements for training AVs, especially in corner cases, like data-missing and limited data scenes. 

\section*{Acknowledgements}
This research is supported by the Science and Technology Development Fund of Macau SAR (File no. 0021/2022/ITP, 0081/2022/A2, 001/2024/SKL), and University of Macau (SRG2023-00037-IOTSC).
\bibliographystyle{named}
\bibliography{ijcai24}

\begin{thebibliography}{}

\bibitem[\protect\citeauthoryear{Alahi \bgroup \em et al.\egroup }{2016}]{alahi2016social}
Alexandre Alahi, Kratarth Goel, Vignesh Ramanathan, Alexandre Robicquet, Li~Fei-Fei, and Silvio Savarese.
\newblock Social lstm: Human trajectory prediction in crowded spaces.
\newblock In {\em Proceedings of the IEEE CVPR}, 2016.

\bibitem[\protect\citeauthoryear{Bhattacharyya \bgroup \em et al.\egroup }{2023}]{bhattacharyya2023ssl}
Prarthana Bhattacharyya, Chengjie Huang, and Krzysztof Czarnecki.
\newblock Ssl-lanes: Self-supervised learning for motion forecasting in autonomous driving.
\newblock In {\em Conference on Robot Learning}, pages 1793--1805. PMLR, 2023.

\bibitem[\protect\citeauthoryear{Boguna \bgroup \em et al.\egroup }{2021}]{boguna2021network}
Marian Boguna, Ivan Bonamassa, Manlio De~Domenico, Shlomo Havlin, Dmitri Krioukov, and M~{\'A}ngeles Serrano.
\newblock Network geometry.
\newblock {\em Nature Reviews Physics}, 3(2):114--135, 2021.

\bibitem[\protect\citeauthoryear{Cai \bgroup \em et al.\egroup }{2021}]{cai2021environment}
Yingfeng Cai, Zihao Wang, Hai Wang, Long Chen, Yicheng Li, Miguel~Angel Sotelo, and Zhixiong Li.
\newblock Environment-attention network for vehicle trajectory prediction.
\newblock {\em IEEE Transactions on Vehicular Technology}, 70(11):11216--11227, 2021.

\bibitem[\protect\citeauthoryear{Chang \bgroup \em et al.\egroup }{2019}]{chang2019argoverse}
Ming-Fang Chang, John Lambert, Patsorn Sangkloy, Jagjeet Singh, Slawomir Bak, Andrew Hartnett, De~Wang, Peter Carr, Simon Lucey, Deva Ramanan, et~al.
\newblock Argoverse: 3d tracking and forecasting with rich maps.
\newblock In {\em Proceedings of the IEEE/CVF CVPR}, 2019.

\bibitem[\protect\citeauthoryear{Chen \bgroup \em et al.\egroup }{2022a}]{chen2022vehicle}
Xiaobo Chen, Huanjia Zhang, Feng Zhao, Yingfeng Cai, Hai Wang, and Qiaolin Ye.
\newblock Vehicle trajectory prediction based on intention-aware non-autoregressive transformer with multi-attention learning for internet of vehicles.
\newblock {\em IEEE Transactions on Instrumentation and Measurement}, 71:1--12, 2022.

\bibitem[\protect\citeauthoryear{Chen \bgroup \em et al.\egroup }{2022b}]{chen2022intention}
Xiaobo Chen, Huanjia Zhang, Feng Zhao, Yu~Hu, Chenkai Tan, and Jian Yang.
\newblock Intention-aware vehicle trajectory prediction based on spatial-temporal dynamic attention network for internet of vehicles.
\newblock {\em IEEE Transactions on Intelligent Transportation Systems}, 23(10):19471--19483, 2022.

\bibitem[\protect\citeauthoryear{Chung \bgroup \em et al.\egroup }{2015}]{chung2015recurrent}
Junyoung Chung, Kyle Kastner, Laurent Dinh, Kratarth Goel, Courville, and Yoshua Bengio.
\newblock A recurrent latent variable model for sequential data.
\newblock {\em Advances in neural information processing systems}, 2015.

\bibitem[\protect\citeauthoryear{Cong \bgroup \em et al.\egroup }{2023}]{cong2023dacr}
Peichao Cong, Yixuan Xiao, Xianquan Wan, Murong Deng, Jiaxing Li, and Xin Zhang.
\newblock Dacr-amtp: Adaptive multi-modal vehicle trajectory prediction for dynamic drivable areas based on collision risk.
\newblock {\em IEEE Transactions on Intelligent Vehicles}, 2023.

\bibitem[\protect\citeauthoryear{Deo and Trivedi}{2018}]{deo2018convolutional}
Nachiket Deo and Mohan~M Trivedi.
\newblock Convolutional social pooling for vehicle trajectory prediction.
\newblock In {\em Proceedings of the IEEE Conference on Computer Vision and Pattern Recognition Workshops}, pages 1468--1476, 2018.

\bibitem[\protect\citeauthoryear{Fang \bgroup \em et al.\egroup }{2020}]{fang2020tpnet}
Liangji Fang, Qinhong Jiang, Jianping Shi, and Bolei Zhou.
\newblock Tpnet: Trajectory proposal network for motion prediction.
\newblock In {\em Proceedings of the IEEE/CVF Conference on Computer Vision and Pattern Recognition}, pages 6797--6806, 2020.

\bibitem[\protect\citeauthoryear{Fernando \bgroup \em et al.\egroup }{2019}]{fernando2019neighbourhood}
Tharindu Fernando, Simon Denman, Sridha Sridharan, and Clinton Fookes.
\newblock Neighbourhood context embeddings in deep inverse reinforcement learning for predicting pedestrian motion over long time horizons.
\newblock In {\em Proceedings of the IEEE/CVF International Conference on Computer Vision Workshops}, pages 0--0, 2019.

\bibitem[\protect\citeauthoryear{Gao \bgroup \em et al.\egroup }{2020}]{gao2020vectornet}
Jiyang Gao, Chen Sun, Hang Zhao, Yi~Shen, Dragomir Anguelov, Congcong Li, and Cordelia Schmid.
\newblock Vectornet: Encoding hd maps and agent dynamics from vectorized representation.
\newblock In {\em Proceedings of the IEEE/CVF Conference on Computer Vision and Pattern Recognition}, pages 11525--11533, 2020.

\bibitem[\protect\citeauthoryear{Gao \bgroup \em et al.\egroup }{2023}]{gao2023dual}
Kai Gao, Xunhao Li, Bin Chen, Lin Hu, Jian Liu, Ronghua Du, and Yongfu Li.
\newblock Dual transformer based prediction for lane change intentions and trajectories in mixed traffic environment.
\newblock {\em IEEE Transactions on Intelligent Transportation Systems}, 2023.

\bibitem[\protect\citeauthoryear{Gilles \bgroup \em et al.\egroup }{2021}]{gilles2021home}
Thomas Gilles, Stefano Sabatini, Dzmitry Tsishkou, Bogdan Stanciulescu, and Fabien Moutarde.
\newblock Home: Heatmap output for future motion estimation.
\newblock In {\em IEEE ITSC}, 2021.

\bibitem[\protect\citeauthoryear{Gilles \bgroup \em et al.\egroup }{2022}]{gilles2022gohome}
Thomas Gilles, Stefano Sabatini, Dzmitry Tsishkou, Bogdan Stanciulescu, and Fabien Moutarde.
\newblock Gohome: Graph-oriented heatmap output for future motion estimation.
\newblock In {\em Proceedings of ICRA}, 2022.

\bibitem[\protect\citeauthoryear{Gu \bgroup \em et al.\egroup }{2021}]{gu2021DenseTNT}
Junru Gu, Chen Sun, and Hang Zhao.
\newblock Densetnt: End-to-end trajectory prediction from dense goal sets.
\newblock In {\em Proceedings of the IEEE/CVF CVPR}, 2021.

\bibitem[\protect\citeauthoryear{Gupta \bgroup \em et al.\egroup }{2018}]{gupta2018social}
Agrim Gupta, Justin Johnson, Li~Fei-Fei, Silvio Savarese, and Alexandre Alahi.
\newblock Social gan: Socially acceptable trajectories with generative adversarial networks.
\newblock In {\em Proceedings of the CVPR}, 2018.

\bibitem[\protect\citeauthoryear{Khandelwal \bgroup \em et al.\egroup }{2020}]{khandelwal2020if}
Siddhesh Khandelwal, William Qi, Jagjeet Singh, Andrew Hartnett, and Deva Ramanan.
\newblock What-if motion prediction for autonomous driving.
\newblock {\em arXiv preprint arXiv:2008.10587}, 2020.

\bibitem[\protect\citeauthoryear{Krajewski \bgroup \em et al.\egroup }{2018}]{highDdataset}
Robert Krajewski, Julian Bock, Laurent Kloeker, and Lutz Eckstein.
\newblock The highd dataset: A drone dataset of naturalistic vehicle trajectories on german highways for validation of highly automated driving systems.
\newblock In {\em Proceeding of the ITSC}, 2018.

\bibitem[\protect\citeauthoryear{Li \bgroup \em et al.\egroup }{2023}]{LiDCHCG23}
Peizheng Li, Shuxiao Ding, Xieyuanli Chen, Niklas Hanselmann, and Juergen Gall.
\newblock Powerbev: {A} powerful yet lightweight framework for instance prediction in bird's-eye view.
\newblock In {\em Proceedings of IJCAI}, 2023.

\bibitem[\protect\citeauthoryear{Liang \bgroup \em et al.\egroup }{2020}]{liang2020learning}
Ming Liang, Bin Yang, Rui Hu, Yun Chen, Renjie Liao, Song Feng, and Raquel Urtasun.
\newblock Learning lane graph representations for motion forecasting.
\newblock In {\em In Proceedings of the ECCV}, 2020.

\bibitem[\protect\citeauthoryear{Liao \bgroup \em et al.\egroup }{2024a}]{10468619}
Haicheng Liao, Yongkang Li, Zhenning Li, Chengyue Wang, Zhiyong Cui, Shengbo~Eben Li, and Chengzhong Xu.
\newblock A cognitive-based trajectory prediction approach for autonomous driving.
\newblock {\em IEEE Transactions on Intelligent Vehicles}, pages 1--12, 2024.

\bibitem[\protect\citeauthoryear{Liao \bgroup \em et al.\egroup }{2024b}]{liao2024bat}
Haicheng Liao, Zhenning Li, Huanming Shen, Wenxuan Zeng, Dongping Liao, Guofa Li, and Chengzhong Xu.
\newblock Bat: Behavior-aware human-like trajectory prediction for autonomous driving.
\newblock In {\em Proceedings of the AAAI Conference on Artificial Intelligence}, volume~38, pages 10332--10340, 2024.

\bibitem[\protect\citeauthoryear{Liao \bgroup \em et al.\egroup }{2024c}]{liao2024cognitive}
Haicheng Liao, Zhenning Li, Chengyue Wang, Bonan Wang, Hanlin Kong, Yanchen Guan, Guofa Li, Zhiyong Cui, and Chengzhong Xu.
\newblock A cognitive-driven trajectory prediction model for autonomous driving in mixed autonomy environment.
\newblock {\em arXiv preprint arXiv:2404.17520}, 2024.

\bibitem[\protect\citeauthoryear{Liao \bgroup \em et al.\egroup }{2024d}]{liao2024human}
Haicheng Liao, Shangqian Liu, Yongkang Li, Zhenning Li, Chengyue Wang, Bonan Wang, Yanchen Guan, and Chengzhong Xu.
\newblock Human observation-inspired trajectory prediction for autonomous driving in mixed-autonomy traffic environments.
\newblock {\em arXiv preprint arXiv:2402.04318}, 2024.

\bibitem[\protect\citeauthoryear{Liao \bgroup \em et al.\egroup }{2024e}]{liao2024gpt}
Haicheng Liao, Huanming Shen, Zhenning Li, Chengyue Wang, Guofa Li, Yiming Bie, and Chengzhong Xu.
\newblock Gpt-4 enhanced multimodal grounding for autonomous driving: Leveraging cross-modal attention with large language models.
\newblock {\em Communications in Transportation Research}, 4:100116, 2024.

\bibitem[\protect\citeauthoryear{Liao \bgroup \em et al.\egroup }{2024f}]{liao2024physics}
Haicheng Liao, Chengyue Wang, Zhenning Li, Yongkang Li, Bonan Wang, Guofa Li, and Chengzhong Xu.
\newblock Physics-informed trajectory prediction for autonomous driving under missing observation.
\newblock {\em Available at SSRN 4809575}, 2024.

\bibitem[\protect\citeauthoryear{Liu \bgroup \em et al.\egroup }{2021}]{liu2021multimodal}
Yicheng Liu, Jinghuai Zhang, Liangji Fang, Qinhong Jiang, and Bolei Zhou.
\newblock Multimodal motion prediction with stacked transformers.
\newblock In {\em Proceedings of the IEEE/CVF CVPR}, 2021.

\bibitem[\protect\citeauthoryear{Markkula \bgroup \em et al.\egroup }{2020}]{markkula2020defining}
Gustav Markkula, Ruth Madigan, Dimitris Nathanael, Evangelia Portouli, Yee~Mun Lee, Andr{\'e} Dietrich, Jac Billington, Anna Schieben, and Natasha Merat.
\newblock Defining interactions: A conceptual framework for understanding interactive behaviour in human and automated road traffic.
\newblock {\em Theoretical Issues in Ergonomics Science}, 21(6):728--752, 2020.

\bibitem[\protect\citeauthoryear{Mercat \bgroup \em et al.\egroup }{2020}]{mercat2020multi}
Jean Mercat, Thomas Gilles, Nicole El~Zoghby, Guillaume Sandou, Dominique Beauvois, and Guillermo~Pita Gil.
\newblock Multi-head attention for multi-modal joint vehicle motion forecasting.
\newblock In {\em IEEE ICRA}, 2020.

\bibitem[\protect\citeauthoryear{Messaoud \bgroup \em et al.\egroup }{2019}]{messaoud2019non}
Kaouther Messaoud, Itheri Yahiaoui, Anne Verroust-Blondet, and Fawzi Nashashibi.
\newblock Non-local social pooling for vehicle trajectory prediction.
\newblock In {\em IEEE Vehicles Symposium}, 2019.

\bibitem[\protect\citeauthoryear{Messaoud \bgroup \em et al.\egroup }{2021}]{messaoud2021attention}
Kaouther Messaoud, Itheri Yahiaoui, Anne Verroust-Blondet, and Fawzi Nashashibi.
\newblock Attention based vehicle trajectory prediction.
\newblock {\em IEEE Transactions on Intelligent Vehicles}, 6(1):175--185, 2021.

\bibitem[\protect\citeauthoryear{Mohamed \bgroup \em et al.\egroup }{2020}]{mohamed2020social}
Abduallah Mohamed, Kun Qian, Mohamed Elhoseiny, and Christian Claudel.
\newblock Social-stgcnn: A social spatio-temporal graph convolutional neural network for human trajectory prediction.
\newblock In {\em Proceedings of the IEEE/CVF CVPR}, 2020.

\bibitem[\protect\citeauthoryear{Ngiam \bgroup \em et al.\egroup }{2022}]{ngiam2022scene}
Jiquan Ngiam, Vijay Vasudevan, Benjamin Caine, Zhengdong Zhang, Hao-Tien~Lewis Chiang, Jeffrey Ling, Rebecca Roelofs, Alex Bewley, Chenxi Liu, Ashish Venugopal, David~J Weiss, Ben Sapp, Zhifeng Chen, and Jonathon Shlens.
\newblock Scene transformer: A unified architecture for predicting future trajectories of multiple agents.
\newblock In {\em Proceedings of the ICLR}, 2022.

\bibitem[\protect\citeauthoryear{Ren \bgroup \em et al.\egroup }{2024}]{ren2024emsin}
Yilong Ren, Zhengxing Lan, Lingshan Liu, and Haiyang Yu.
\newblock Emsin: Enhanced multi-stream interaction network for vehicle trajectory prediction.
\newblock {\em IEEE Transactions on Fuzzy Systems}, 2024.

\bibitem[\protect\citeauthoryear{Schwarting \bgroup \em et al.\egroup }{2019}]{schwarting2019social}
Wilko Schwarting, Alyssa Pierson, Javier Alonso-Mora, Sertac Karaman, and Daniela Rus.
\newblock Social behavior for autonomous vehicles.
\newblock {\em Proceedings of the National Academy of Sciences}, 116:24972, 2019.

\bibitem[\protect\citeauthoryear{Song \bgroup \em et al.\egroup }{2022}]{song2022learning}
Haoran Song, Di~Luan, Wenchao Ding, Michael~Y Wang, and Qifeng Chen.
\newblock Learning to predict vehicle trajectories with model-based planning.
\newblock In {\em Conference on Robot Learning}, pages 1035--1045. PMLR, 2022.

\bibitem[\protect\citeauthoryear{Tian \bgroup \em et al.\egroup }{2024}]{tian2024hydralora}
Chunlin Tian, Zhan Shi, Zhijiang Guo, Li~Li, and Chengzhong Xu.
\newblock Hydralora: An asymmetric lora architecture for efficient fine-tuning, 2024.

\bibitem[\protect\citeauthoryear{Toghi \bgroup \em et al.\egroup }{2022}]{toghi2022social}
Behrad Toghi, Rodolfo Valiente, Dorsa Sadigh, Ramtin Pedarsani, and Yaser~P Fallah.
\newblock Social coordination and altruism in autonomous driving.
\newblock {\em IEEE Transactions on Intelligent Transportation Systems}, 2022.

\bibitem[\protect\citeauthoryear{Walters \bgroup \em et al.\egroup }{2021}]{walters2021trajectory}
Robin Walters, Jinxi Li, and Rose Yu.
\newblock Trajectory prediction using equivariant continuous convolution.
\newblock In {\em Proceedings of the ICLR}, 2021.

\bibitem[\protect\citeauthoryear{Wang \bgroup \em et al.\egroup }{2020}]{wang2020linformer}
Sinong Wang, Belinda~Z Li, Madian Khabsa, Han Fang, and Hao Ma.
\newblock Linformer: Self-attention with linear complexity.
\newblock {\em arXiv preprint:2006.04768}, 2020.

\bibitem[\protect\citeauthoryear{Wang \bgroup \em et al.\egroup }{2022a}]{wang2022ltp}
Jingke Wang, Tengju Ye, Ziqing Gu, and Junbo Chen.
\newblock Ltp: Lane-based trajectory prediction for autonomous driving.
\newblock In {\em Proceedings of the IEEE/CVF Conference on Computer Vision and Pattern Recognition}, pages 17134--17142, 2022.

\bibitem[\protect\citeauthoryear{Wang \bgroup \em et al.\egroup }{2022b}]{wang2022multi}
Yu~Wang, Shengjie Zhao, Rongqing Zhang, Xiang Cheng, and Liuqing Yang.
\newblock Multi-vehicle collaborative learning for trajectory prediction with spatio-temporal tensor fusion.
\newblock {\em IEEE Transactions on Intelligent Transportation Systems}, 23(1):236--248, 2022.

\bibitem[\protect\citeauthoryear{Wang \bgroup \em et al.\egroup }{2023}]{wang2023wsip}
Renzhi Wang, Senzhang Wang, Hao Yan, and Xiang Wang.
\newblock Wsip: Wave superposition inspired pooling for dynamic interactions-aware trajectory prediction.
\newblock In {\em Proceedings of the AAAI Conference on Artificial Intelligence}, volume~37, pages 4685--4692, 2023.

\bibitem[\protect\citeauthoryear{Wu and He}{2018}]{wu2018group}
Yuxin Wu and Kaiming He.
\newblock Group normalization.
\newblock In {\em Proceedings of the ECCV}, 2018.

\bibitem[\protect\citeauthoryear{Xie \bgroup \em et al.\egroup }{2021}]{xie2021congestion}
Xu~Xie, Chi Zhang, Yixin Zhu, Ying~Nian Wu, and Song-Chun Zhu.
\newblock Congestion-aware multi-agent trajectory prediction for collision avoidance.
\newblock In {\em Proceedings of the IEEE ICRA}, 2021.

\bibitem[\protect\citeauthoryear{Xue \bgroup \em et al.\egroup }{2021}]{xue2021hierarchical}
Qifan Xue, Shengyi Li, Xuanpeng Li, Jingwen Zhao, and Weigong Zhang.
\newblock Hierarchical motion encoder-decoder network for trajectory forecasting.
\newblock {\em arXiv preprint arXiv:2111.13324}, 2021.

\bibitem[\protect\citeauthoryear{Ye \bgroup \em et al.\egroup }{2021}]{ye2021tpcn}
Maosheng Ye, Tongyi Cao, and Qifeng Chen.
\newblock Tpcn: Temporal point cloud networks for motion forecasting.
\newblock In {\em Proceedings of the CVPR}, 2021.

\bibitem[\protect\citeauthoryear{Yin \bgroup \em et al.\egroup }{2021}]{yin2021multimodal}
Ziyi Yin, Ruijin Liu, Zhiliang Xiong, and Zejian Yuan.
\newblock Multimodal transformer networks for pedestrian trajectory prediction.
\newblock In {\em IJCAI}, 2021.

\bibitem[\protect\citeauthoryear{Zeng \bgroup \em et al.\egroup }{2021}]{zeng2021lanercnn}
Wenyuan Zeng, Ming Liang, Renjie Liao, and Raquel Urtasun.
\newblock Lanercnn: Distributed representations for graph-centric motion forecasting.
\newblock In {\em IEEE/RSJ IROS}, 2021.

\bibitem[\protect\citeauthoryear{Zhao \bgroup \em et al.\egroup }{2021}]{zhao2021tnt}
Hang Zhao, Jiyang Gao, Tian Lan, Chen Sun, Ben Sapp, Balakrishnan Varadarajan, Yue Shen, Yi~Shen, Yuning Chai, Cordelia Schmid, et~al.
\newblock Tnt: Target-driven trajectory prediction.
\newblock In {\em Conference on Robot Learning}, pages 895--904. PMLR, 2021.

\bibitem[\protect\citeauthoryear{Zhou \bgroup \em et al.\egroup }{2022}]{Zhou_2022_CVPR}
Zikang Zhou, Luyao Ye, Jianping Wang, Kui Wu, and Kejie Lu.
\newblock Hivt: Hierarchical vector transformer for multi-agent motion prediction.
\newblock In {\em Proceedings of the IEEE/CVF CVPR}, 2022.

\bibitem[\protect\citeauthoryear{Zhou \bgroup \em et al.\egroup }{2023}]{zhou2023query}
Zikang Zhou, Jianping Wang, Yung-Hui Li, and Yu-Kai Huang.
\newblock Query-centric trajectory prediction.
\newblock In {\em Proceedings of the IEEE/CVF CVPR}, 2023.

\bibitem[\protect\citeauthoryear{Zuo \bgroup \em et al.\egroup }{2023}]{zuo2023trajectory}
Zhiqiang Zuo, Xinyu Wang, Songlin Guo, Zhengxuan Liu, Zheng Li, and Yijing Wang.
\newblock Trajectory prediction network of autonomous vehicles with fusion of historical interactive features.
\newblock {\em IEEE Transactions on Intelligent Vehicles}, 2023.

\end{thebibliography}

\end{document}